\documentclass[10pt,twocolumn,letterpaper]{article}

\usepackage[pagenumbers]{cvpr} %

\usepackage{graphicx}
\usepackage{amsmath}
\usepackage{amssymb}
\usepackage{booktabs}

\usepackage{makecell}
\usepackage{multirow}

\usepackage[pagebackref,breaklinks,colorlinks]{hyperref}

\usepackage[capitalize]{cleveref}
\crefname{section}{Sec.}{Secs.}
\Crefname{section}{Section}{Sections}
\Crefname{table}{Table}{Tables}
\crefname{table}{Tab.}{Tabs.}

\def\cvprPaperID{***} %
\def\confName{CVPR}
\def\confYear{2023}

\newcommand{\ignorethis}[1]{}

\begin{document}

\title{Tri-Perspective View for Vision-Based 3D Semantic Occupancy Prediction}

\newcommand*\samethanks[1][\value{footnote}]{\footnotemark[#1]}
\author{Yuanhui Huang$^{1,2,}$\footnotetext{sadfsadfa}\thanks{Equal contribution.}\quad Wenzhao Zheng$^{1,2,}$\samethanks\quad Yunpeng Zhang$^3$\quad Jie Zhou$^{1,2}$\quad Jiwen Lu$^{1,2,}$\thanks{Corresponding author.} \\
$^1$Beijing National Research Center for Information Science and Technology, China \\
$^2$Department of Automation, Tsinghua University, China \quad\quad $^3$PhiGent Robotics \\
\texttt{\{huangyh22,zhengwz18\}@mails.tsinghua.edu.cn; yunpengzhang97@gmail.com;} \\
\texttt{\{jzhou,lujiwen\}@tsinghua.edu.cn}
}

\newcommand\blfootnote[1]{%
  \begingroup
  \renewcommand\thefootnote{}\footnote{#1}%
  \addtocounter{footnote}{-1}%
  \endgroup
}

\twocolumn[{%
\renewcommand\twocolumn[1][]{#1}%
\maketitle
\vspace{-12mm}
\begin{center}
    \centering
    \includegraphics[width=\linewidth]{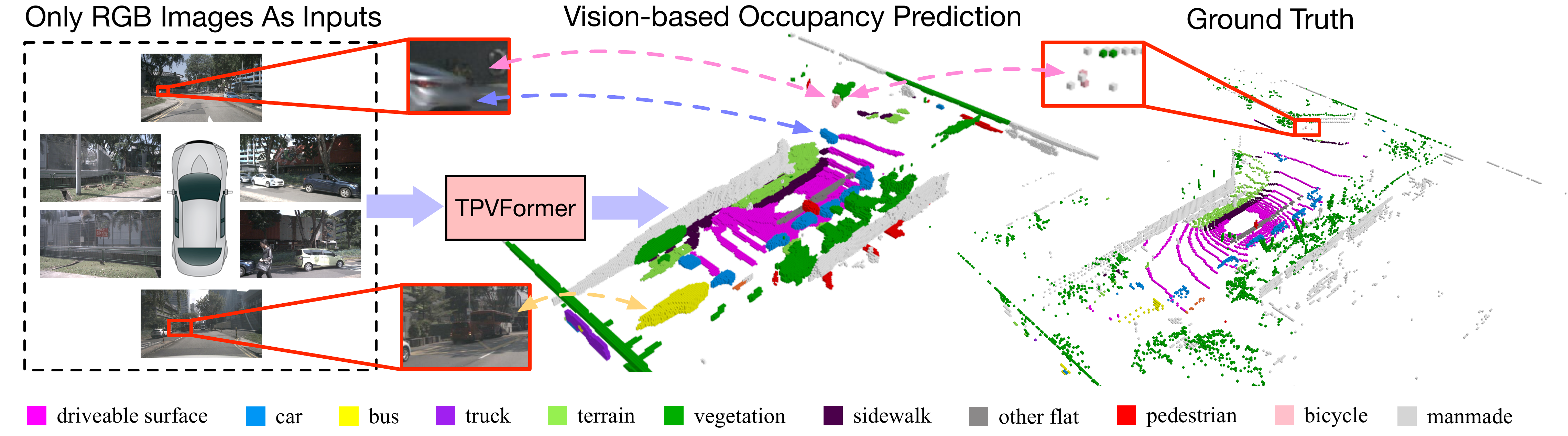}
    \vspace{-7mm}
    \captionof{figure}{
    Given only surround-camera RGB images as inputs, our model (trained using only sparse LiDAR point supervision) can predict the semantic occupancy for all volumes in the 3D space.
    This task is challenging as it requires both geometric and semantic understandings of the 3D scene.
    We observe that our model can produce even more comprehensive and consistent volume occupancy than the groundtruth on the validation set (not seen during training) of nuScenes~\cite{nuscenes}.
    Despite the lack of geometric inputs like LiDAR, our model can accurately identify the 3D positions and sizes of close and distant objects.
    Particularly, our model even successfully identifies the partially occluded bicycle captured only by two LiDAR points, demonstrating the potential advantage of vision-based 3D semantic occupancy prediction.
    }
\label{teaser}
\end{center}%
}]

\begin{abstract}
Modern methods for vision-centric autonomous driving perception widely adopt the bird's-eye-view (BEV) representation to describe a 3D scene.
Despite its better efficiency than voxel representation, it has difficulty describing the fine-grained 3D structure of a scene with a single plane.
To address this, we propose a tri-perspective view (TPV) representation which accompanies BEV with two additional perpendicular planes.
We model each point in the 3D space by summing its projected features on the three planes. 
To lift image features to the 3D TPV space, we further propose a transformer-based TPV encoder (TPVFormer) to obtain the TPV features effectively.
We employ the attention mechanism to aggregate the image features corresponding to each query in each TPV plane. 
Experiments show that our model trained with sparse supervision effectively predicts the semantic occupancy for all voxels.
We demonstrate for the first time that using only camera inputs can achieve comparable performance with LiDAR-based methods on the LiDAR segmentation task on nuScenes.
Code: \url{https://github.com/wzzheng/TPVFormer}.

\end{abstract}

\section{Introduction}
Perceiving the 3D surroundings accurately and comprehensively plays an important role in the autonomous driving system.
Vision-based 3D perception recently emerges as a promising alternative to LiDAR-based one to effectively extract 3D information from 2D images.
Though lacking direct sensing of depth information, vision-based models empowered by surrounding cameras demonstrate promising performance on various 3D perception tasks such as depth estimation~\cite{guizilini2021full,wei2022surround}, semantic map reconstruction~\cite{beverse,fiery,stretchbev}, and 3D object detection~\cite{petr,bevformer,simmod}.
\blfootnote{*Equal contribution. $\dagger$Corresponding author.}

The core of 3D surrounding perceiving lies in how to effectively represent a 3D scene.
Conventional methods split the 3D space into voxels and assign each voxel a vector to represent its status.
Despite its accuracy, the vast number of voxels poses a great challenge to computation and requires specialized techniques like sparse convolution~\cite{choy20194d}.
As the information in outdoor scenes is not isotropically distributed, modern methods collapse the height dimension and mainly focus on the ground plane (bird's-eye-view) where information varies the most~\cite{bevdepth,bevfusion,bevfusion2,lss,bevdet,beverse,simmod}.
They implicitly encode the 3D information of each object in the vector representation in each BEV grid.
Though more efficient, BEV-based methods perform surprisingly well on the 3D object detection task~\cite{bevfusion,bevfusion2}.
This is because 3D object detection only demands predictions of coarse-level bounding boxes for commonly seen objects such as cars and pedestrians.
However, objects with various 3D structures can be encountered in real scenes and it is difficult (if not impossible) to encode all of them using a flattened vector.
Therefore, it requires a more comprehensive and fine-grained understanding of the 3D surroundings toward a safer and more robust vision-centric autonomous driving system.
Still, it remains unknown how to generalize BEV to model fine-grained 3D structures while preserving its efficiency and detection performance.

\begin{figure}[t]
\centering
\includegraphics[width=0.475\textwidth]{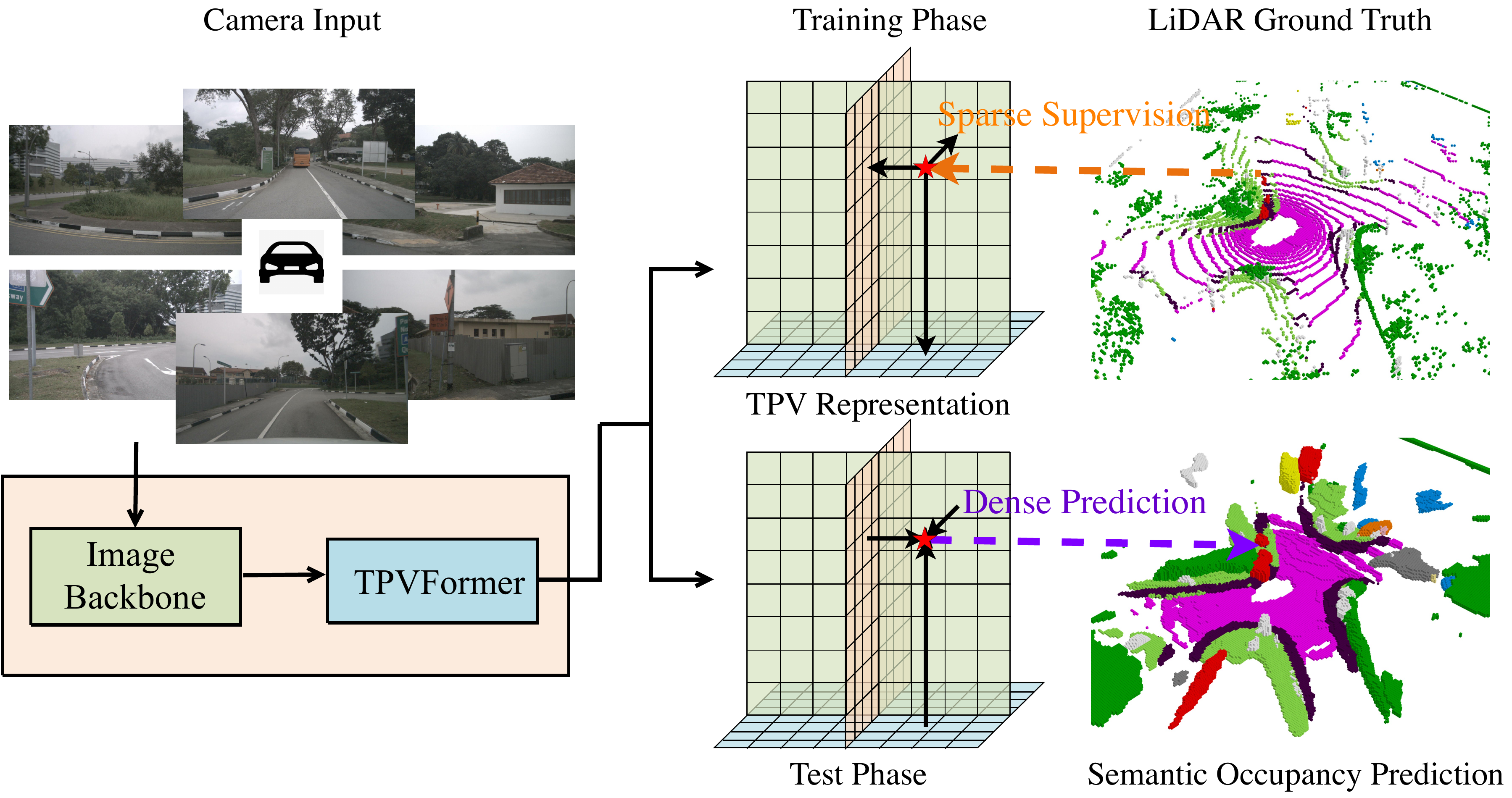}
\vspace{-7mm}
\caption{An overview of our method for 3D semantic occupancy prediction.
Taking camera images as inputs, the proposed TPVFormer only uses sparse LiDAR semantic labels for training but can effectively predict the semantic occupancy for all voxels.
}
\label{fig:overview}
\vspace{-7mm}
\end{figure}

In this paper, we advance in this direction and propose a tri-perspective view (TPV) representation to describe a 3D scene.
Motivated by recent advances in explicit-implicit hybrid scene representations~\cite{triplane,tensorf}, we generalize BEV by accompanying it with two perpendicular planes to construct three cross-planes perpendicular to each other.
Each plane models the 3D surroundings from one view and combining them provides a comprehensive description of the 3D structure.
Specifically, to obtain the feature of a point in the 3D space, we first project it into each of the three planes and use bilinear interpolation to obtain the feature for each projected point.
We then sum the three projected features as the comprehensive feature of the 3D point.
The TPV representation is thus able to describe the 3D scene at an arbitrary resolution and produces different features for different points in the 3D space.
We further propose a transformer-based encoder (TPVFormer) to effectively obtain the TPV features from 2D images.
We first perform image cross-attention between TPV grid queries and the corresponding 2D image features to lift 2D information to the 3D space.
We then perform cross-view hybrid-attention among the TPV features to enable interactions among the three planes.

To demonstrate the superiority of TPV, we formulate a practical yet challenging task for vision-based 3D semantic occupancy prediction, where only sparse lidar semantic labels are provided for training and predictions for all voxels are required for testing, as shown in Figure~\ref{fig:overview}.
However, as no benchmark is provided on this challenging setting, we only perform qualitative analysis but provide a quantitative evaluation on two proxy tasks: LiDAR segmentation (sparse training, sparse testing) on nuScenes~\cite{nuscenes} and 3D semantic scene completion (dense training, dense testing) on SemanticKITTI~\cite{semantickitti}.
For both tasks, we only use RGB images as inputs.
For LiDAR segmentation, our model use the LiDAR data only for point query to compute evaluation metrics.
Visualization results show that TPVFormer produces consistent semantic voxel occupancy prediction with only sparse point supervision during training, as shown in Figure~\ref{teaser}.
We also demonstrate for the first time that our vision-based method achieves comparable performance with LiDAR-based methods on LiDAR segmentation.

\section{Related Work}

\textbf{Voxel-based Scene Representation:}
Obtaining an effective representation for a 3D scene is the basic procedure for 3D surrounding perception.
One direct way is to discretize the 3D space into voxels and assign a vector to represent each voxel~\cite{voxelnet,zhu2021cylindrical}.
The ability to describe fine-grained 3D structures makes voxel-based representation favorable for 3D semantic occupancy prediction tasks including lidar segmentation~\cite{amvnet,spvnas,af2s3net,cylinder3D,drinet++,ye2022lidarmultinet} and 3D scene completion~\cite{monoscene,lmscnet,dsketch,aicnet,js3c}.
Though they have dominated the 3D segmentation task~\cite{ye2022lidarmultinet}, they still lag behind BEV-based methods on the 3D detection performance~\cite{bevdepth}.
Despite the success of voxel-based representations in LiDAR-centric surrounding perception, only a few works have explored voxel-based representations for vision-centric autonomous driving~\cite{uvtr,monoscene}.
MonoScene~\cite{monoscene} first backprojects image features to all possible positions in the 3D space along the optical ray to obtain the initial voxel representation and further processes it using a 3D UNet.
However, it is still challenging to generalize it to 3D perception with multi-view images due to the inefficiency of voxel representations.
This motivates us to explore more efficient and expressive ways to describe the fine-grained 3D structure of a scene.

\textbf{BEV-based Scene Representation:}
The vast number of voxels poses a great challenge to the computation efficiency of voxel-based methods.
Considering that the height dimension contains less information than the other two dimensions, BEV-based methods implicitly encode the height information in each BEV grid for a more compact representation of scenes~\cite{pointpillars}.
Recent studies in BEV-based perception focus on how to effectively transform features from the image space to the BEV space~\cite{caddn,bevdepth,lss,bevdet,beverse,bevformer}.
One line of works explicitly predict a depth map for each image and utilizes it to project image features into the 3D space followed by BEV pooling~\cite{caddn,bevdepth,bevfusion,bevfusion2,lss,bevdet,beverse}.
Another line of works employ BEV queries to implicitly assimilate information from image features using the cross-attention mechanism~\cite{bevformer,polarformer}.
BEV-based perception achieves great success on vision-centric 3D detection from multi-view images~\cite{bevdepth}, demonstrating comparable performance to LiDAR-centric methods.
Yet, it is difficult to apply BEV to 3D semantic occupancy prediction which requires a more fine-grained description of the 3D space.

\textbf{Implicit Scene Representation:}
Recent methods have also explored implicit representations to describe a scene.
They learn a continuous function that takes as input the 3D coordinate of a point and outputs the representation of this point~\cite{mescheder2019occupancy,nerf,deepsdf}.
Compared with explicit representations like voxel and BEV, implicit representations usually share the advantage of arbitrary-resolution modeling and computation-efficient architectures~\cite{chabra2020deep,chen2021learning,reiser2021kilonerf}.
These advantages enable them to scale to larger and more complex scenes with more fine-grained descriptions. 
Especially, our work is inspired by recent advances in hybrid explicit-implicit representations~\cite{triplane,tensorf}.
They explicitly inject spatial information into the continuous mapping of implicit representations.
Therefore, they share the computation-efficient architecture of implicit representations and better spatial awareness of explicit representations.
Still, they mainly focus on small-scale complex scenes for 3D-aware image rendering.
To the best of our knowledge, we are the first to use implicit representation to model outdoor scenes for 3D surrounding perception in autonomous driving.
\section{Proposed Approach}

\subsection{Generalizing BEV to TPV}
Autonomous driving perception typically requires both expressive and efficient representation of the complex 3D scene, among which voxel and Bird's-Eye-View (BEV) representations are the two most widely adopted frameworks.
Voxel representation~\cite{drinet++,spvnas,uvtr} describes a 3D scene with dense cubic features $\mathbf{V}\in\mathbb{R}^{H\times W\times D\times C}$ where $H$, $W$, $D$ are the spatial resolution of the voxel space and $C$ denotes the feature dimension.
A random point located at $(x, y, z)$ in the real world maps to its voxel coordinates $(h, w, d)$ through one-to-one correspondence $\mathcal{P}_{vox}$, and the resulting feature $\mathbf{f}_{x,y,z}$ is obtained by sampling $\mathbf{V}$ at $(h, w, d)$:
\begin{equation}\label{eqn: BEV proj}
\begin{aligned}
    \mathbf{f}_{x,y,z}=\mathbf{v}_{h,w,d}&=\mathcal{S}(\mathbf{V}, (h,w,d)), \\
    &=\mathcal{S}(\mathbf{V}, \mathcal{P}_{vox}(x,y,z)),
\end{aligned}
\end{equation}
where $\mathcal{S}(arg1, arg2)$ denotes sampling $arg1$ at the position specified in $arg2$ and $\mathbf{v}_{h,w,d}$ is the sampled voxel feature.
Note that the projection function $\mathcal{P}_{vox}$ is composed of simple scaling and rigid transformations if the voxel space aligns with the real world.
Therefore, voxel representation preserves the dimensionality of the real world and offers sufficient expressiveness with appropriate $H,W,D$.
Yet, the storage and computation complexity of voxel features are proportion to $O(HWD)$, making it challenging to deploy them in real-time onboard applications.

As a popular alternative, BEV~\cite{polarformer,bevformer,bevdepth,bevfusion} models the 3D scene with a 2D feature map $\mathbf{B}\in\mathbb{R}^{H\times W\times C}$ which encodes the top view of the scene.
Different from the voxel counterpart, the point at $(x, y, z)$ is projected to its BEV coordinates $(h, w)$ using only the positional information from the ground plane regardless of the $z$-axis.
Each feature $\mathbf{b}_{h, w}$ sampled from $\mathbf{B}$ corresponds to a pillar region covering the full range of $z$-axis in the real world:
\begin{equation}
    \begin{aligned}
        \mathbf{f}_{x,y,\mathbf{Z}}=\mathbf{b}_{h,w}=\mathcal{S}(\mathbf{B}, (h,w))=\mathcal{S}(\mathbf{B}, \mathcal{P}_{bev}(x,y)),
    \end{aligned}
\end{equation}
where $\mathbf{f}_{x,y,\mathbf{Z}}$ denotes features of points sharing the same $(x,y)$ but differing in $z$, and $\mathcal{P}_{bev}$ is the point-to-BEV projection.
Although BEV greatly reduces the storage and computation burden to $O(HW)$, completely omitting the $z$-axis has an adverse effect on its expressiveness. %

\begin{figure}[t]
\centering
\includegraphics[width=0.475\textwidth]{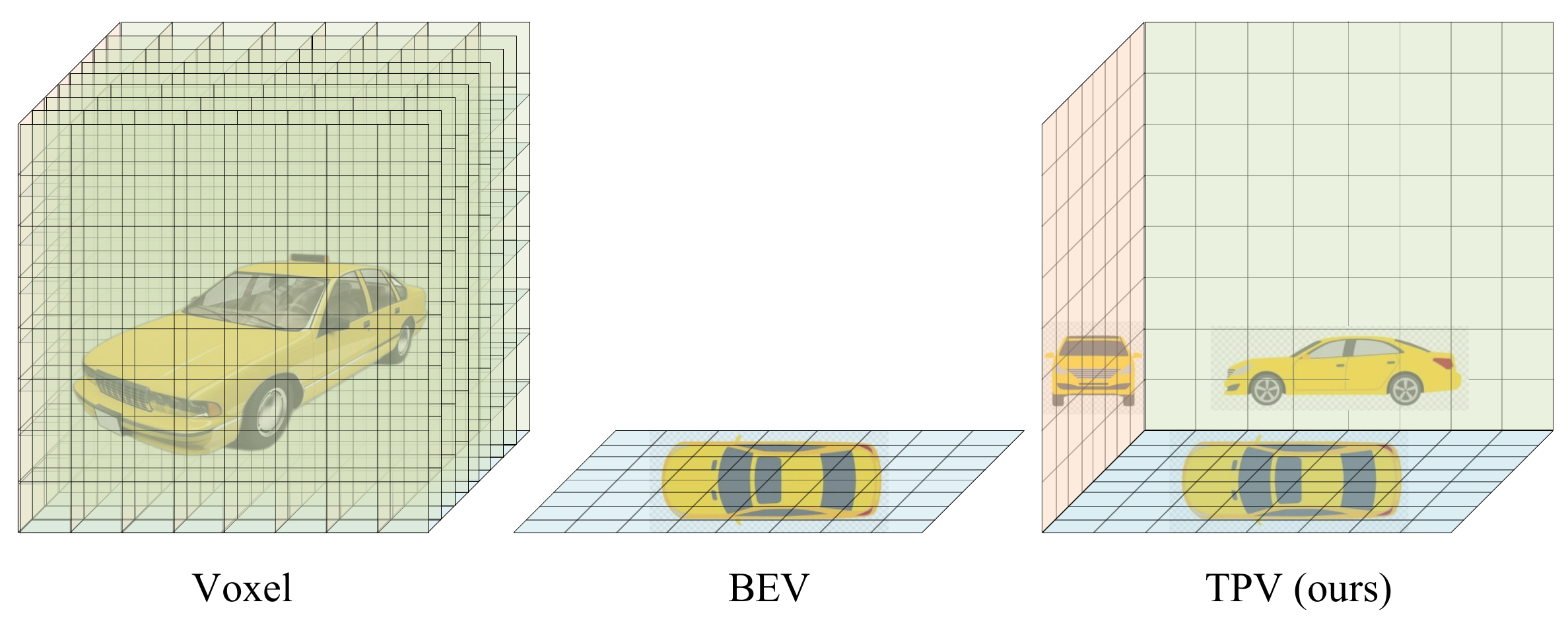}
\vspace{-7mm}
\caption{Comparisons of the proposed TPV representation with voxel and BEV representation.
While BEV is more efficient than the voxel representation, it discards the height information and cannot comprehensively describe a 3D scene.
}
\label{fig:comparisons}
\vspace{-6mm}
\end{figure}

To address this, we propose a Tri-Perspective View (TPV) representation which is capable of modeling the 3D space at full scale without suppressing any axes and avoiding cubic complexity, as illustrated in Figure~\ref{fig:comparisons}.
Formally, we learn three axis-aligned orthogonal TPV planes:
\begin{equation}\label{eqn: tpv planes def}
\begin{aligned}
    &\mathbf{T} = [\mathbf{T}^{HW}, \mathbf{T}^{DH}, \mathbf{T}^{WD}], \ \  \mathbf{T}^{HW}\in\mathbb{R}^{H\times W\times C}, \\
    &\mathbf{T}^{DH}\in\mathbb{R}^{D\times H\times C}, \ \  \mathbf{T}^{WD}\in\mathbb{R}^{W\times D\times C},
\end{aligned}
\end{equation}
which represent the top, side and front views of a 3D scene respectively.
$H,W,D$ denote the resolution of the three planes and $C$ is the feature dimension.
Intuitively, a complex scene, when examined from different perspectives, can be better understood because these perspectives may provide complementary clues about the scene.

\textbf{Point Querying formulation.}
Given a query point at $(x,y,z)$ in the real world, TPV representation tries to aggregate its projections on the top, side and front views in order to get a comprehensive description of the point.
To elaborate, we first project the point onto the TPV planes to obtain the coordinates $[(h,w),(d,h),(w,d)]$, sample the TPV planes at these locations to retrieve the corresponding features $[\mathbf{t}_{h,w},\mathbf{t}_{d,h},\mathbf{t}_{w,d}]$, and aggregate the three features to generate the final $\mathbf{f}_{x,y,z}$:
\begin{equation}\label{eqn: tpv plane sampling}
    \begin{aligned}
        \mathbf{t}_{h,w} &= \mathcal{S}(\mathbf{T}^{HW},(h,w)) = \mathcal{S}(\mathbf{T}^{HW},\mathcal{P}_{hw}(x,y)), \\
        \mathbf{t}_{d,h} &= \mathcal{S}(\mathbf{T}^{DH},(d,h)) = \mathcal{S}(\mathbf{T}^{DH},\mathcal{P}_{dh}(z,x)), \\
        \mathbf{t}_{w,d} &= \mathcal{S}(\mathbf{T}^{WD},(w,d)) = \mathcal{S}(\mathbf{T}^{WD},\mathcal{P}_{wd}(y,z)).
    \end{aligned}
\end{equation}
\begin{equation}\label{eqn: tpv plane aggregate}
    \begin{aligned}
        \mathbf{f}_{x,y,z}=\mathcal{A}(\mathbf{t}_{h,w}, \mathbf{t}_{d,h}, \mathbf{t}_{w,d}),
    \end{aligned}
\end{equation}
where the sampling function $\mathcal{S}$ and the aggregation function $\mathcal{A}$ are implemented with bilinear interpolation and summation respectively, and each projection function $\mathcal{P}$ performs simple scaling on the two relevant coordinates since the TPV planes are aligned with the real-world axes.

\textbf{Voxel feature formulation.}
Equivalent to the point querying formulation, the TPV planes, when expanded along respective orthogonal directions and summed up, construct a full-scale 3D feature space similar to the voxel feature space, but only with storage and computation complexity of $O(HW+DH+WD)$, which is an order of magnitude lower than the voxel counterpart.

Compared with BEV, as the three planes in TPV are perpendicular to each other, point features along the orthogonal direction of one plane are diversified by features sampled from the other two planes, which is ignored by the BEV representation.
Moreover, a grid feature in each TPV plane is only responsible for view-specific information of the corresponding pillar region rather than encoding the complete information as in BEV.
To sum up, TPV representation generalizes BEV from single top view to complementary and orthogonal top, side and front views and is able to offer a more comprehensive and fine-grained understanding of the 3D surroundings while remaining efficient.

\begin{figure*}[t]
\centering
\includegraphics[width=\textwidth]{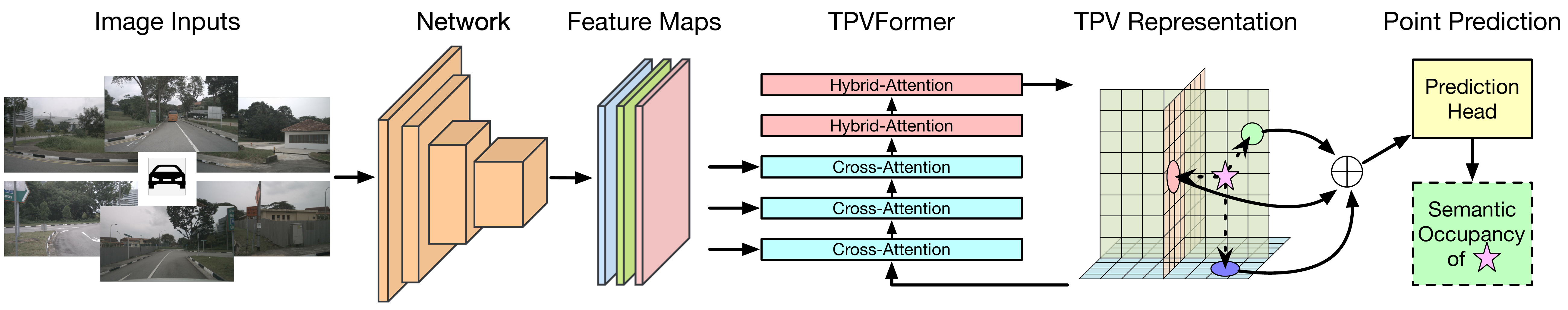}
\vspace{-7mm}
\caption{Framework of the proposed TPVFormer for 3D semantic occupancy prediction.
We employ an image backbone network to extract multi-scale features for multi-camera images. 
We then perform cross-attention to adaptively lift 2D features to the TPV space and use cross-view hybrid attention to enable the interactions between TPV planes.
To predict the semantic occupancy of a point in the 3D space, we apply a lightweight prediction head on the sum of projected features on the three TPV planes.
}
\label{fig:framework}
\vspace{-6mm}
\end{figure*}

\subsection{TPVFormer}
For vision-centric autonomous driving perception, a 2D backbone is often employed to obtain image features before feeding them into a specific encoder depending on the representation framework.
We present a transformer-based TPV encoder (TPVFormer) to lift the image features to the TPV planes through the attention mechanism.

\textbf{Overall Structure:}
In TPVFormer, we introduce TPV queries, image cross-attention (ICA) and cross-view hybrid-attention (CVHA) to enable effective generation of TPV planes, as shown in Fig.~\ref{fig:framework}.
In fact, TPV queries and TPV planes refer to the same set of feature vectors defined in \eqref{eqn: tpv planes def}.
Each TPV query $\mathbf{t}\in\mathbf{T}$ is a grid cell feature belonging to one of the three planes and used to encode view-specific information from the corresponding pillar region.
Cross-view hybrid-attention enables direct interactions among TPV queries from the same or different views in order to gather contextual information.
Inside image cross-attention, TPV queries aggregate visual information from image features through deformable attention.

We further construct two kinds of transformer blocks: hybrid-cross-attention block (HCAB) and hybrid-attention block (HAB).
Composed of both CVHA and ICA attention, the HCAB block is employed in the first half of TPVFormer to effectively query visual information from image features.
Following HCAB blocks, the HAB block contains only CVHA attention and specializes in contextual information encoding. 
Finally, we build TPVFormer by stacking $N_1$ HCAB blocks and $N_2$ HAB blocks.

\textbf{TPV Queries:}
Although TPV queries and TPV planes refer to the same list of 2D features defined in \eqref{eqn: tpv planes def}, they are used in attention and 3D representation contexts, respectively.
Each TPV query maps to a 2D grid cell region of size $s\times s\ m^2$ in the corresponding view, and further to a 3D pillar region extending from the view in the perpendicular direction.
In our pipeline, TPV queries are first enhanced with raw visual information from image features in HCAB blocks, and then refined with contextual clues from other queries in HAB blocks.
As for implementation, we initialize TPV queries as learnable parameters and add 3D positional embedding to them before the first encoder layer.

\textbf{Image Cross-Attention:}
In TPVFormer, we use image cross-attention to lift multi-scale and possibly multi-camera image features to the TPV planes.
Considering the high resolution nature of TPV queries ($\sim 10^4$ queries) and multiple image feature maps ($\sim 10^5$ pixels each), it is unfeasible to compute full-scale vanilla cross-attention between them.
As a workaround, we employ the efficient deformable attention~\cite{deformable_DETR,dcn} to implement image cross-attention.

We take the local receptive field as an inductive bias when sampling the reference points.
Specifically, for a TPV query $\mathbf{t}_{h,w}$ located at $(h,w)$ in the top plane, we first calculate its coordinates $(x,y)$ in the top view in the real world through the inverse projection function $\mathcal{P}_{hw}^{-1}$.
Then we sample uniformly $N_{ref}^{HW}$ reference points for the query $\mathbf{t}_{h,w}$ along the orthogonal direction of the plane:
\begin{equation}
    (x,y) = \mathcal{P}_{hw}^{-1}(h,w) = ((h-\frac{H}{2})\times s, (w-\frac{W}{2})\times s).
\end{equation}
\begin{equation}
    \mathbf{Ref}_{h,w}^{world} = (\mathcal{P}_{hw}^{-1}(h,w), \mathbf{Z}) = \{(x, y, z_{i})\}_{i=1}^{N_{ref}^{HW}},
\end{equation}
where $\mathbf{Ref}_{h,w}^{world}$ denotes the set of reference points in the world coordinate for query $\mathbf{t}_{h,w}$.
The similar procedure is repeated for all TPV queries, and note that the number of reference points $N_{ref}$ may change across planes because of the different ranges of axes. 
After deriving the reference points for $\mathbf{t}_{h,w}$, we need to project them into the pixel coordinate in order to sample the image feature maps later:
\begin{equation}
    \mathbf{Ref}_{h,w}^{pix} = \mathcal{P}_{pix}(\mathbf{Ref}_{h,w}^{world}) = \mathcal{P}_{pix}(\{(x, y, z_{i})\}),
\end{equation}
where $\mathbf{Ref}_{h,w}^{pix}$ is the set of reference points in the pixel coordinate for query $\mathbf{t}_{h,w}$ and $\mathcal{P}_{pix}$ is the perspective projection function determined by the camera extrinsic and intrinsic.
Note that we may have multiple cameras in different directions which will generate a set of $\{\mathbf{Ref}_{h,w}^{pix,j}\}_{j=1}^{N_{c}}$ where $N_{c}$ denotes the number of cameras.
Since not all cameras can capture the reference points of query $\mathbf{t}_{h,w}$, we can further reduce computation by removing invalid sets from $\{\mathbf{Ref}_{h,w}^{pix,j}\}_{j=1}^{N_{c}}$ if none of the reference points falls onto the image captured by the corresponding camera.
The final step is to generate offsets and attention weights through two linear layers applied on $\mathbf{t}_{h,w}$ and produce the updated TPV queries by summing up the sampled image features weighted by their attention weights:
\begin{small}
\begin{equation}
    \mathrm{ICA}(\mathbf{t}_{h,w}, \mathbf{I}) \!=\! \frac{1}{|N_{h,w}^{val}|}\!\sum_{j\in N_{h,w}^{val}}\!\!\mathrm{DA}(\mathbf{t}_{h,w}, \mathbf{Ref}_{h,w}^{pix,j}, \mathbf{I}_{j}),
\end{equation}
\end{small}
where $N_{h,w}^{val}$, $\mathbf{I}_{j}$, $\mathrm{DA}(\cdot)$ denote the index set of valid cameras, the image features from the $j$th camera and the deformable attention function, respectively.

\textbf{Cross-View Hybrid-Attention:}
In image cross-attention, TPV queries sample reference image features separately and no direct interactions between them are enabled.
Therefore, we propose cross-view hybrid-attention to allow queries to exchange their information across different views, which benefits context extraction.
We also adopt deformable attention here to reduce computation, in which three TPV planes serve as key and value.
Taking the TPV query $\mathbf{t}_{h,w}$ located at $(h,w)$ in the top plane as an example, we group its reference points into three disjoint subsets, which contains reference points belonging to the top, side and front planes respectively:
\begin{equation}
    \mathbf{R}_{h,w} = \mathbf{R}_{h,w}^{top} \cup \mathbf{R}_{h,w}^{side} \cup \mathbf{R}_{h,w}^{front}.
\end{equation}
To collect reference points on the top plane, we simply sample a few random points in the neighborhood of the query $\mathbf{t}_{h,w}$.
As for the side and front planes, we first sample 3D points uniformly along the direction perpendicular to the top plane and project them onto the side and front planes:
\begin{equation}
    \mathbf{R}_{h,w}^{side} = \{(d_{i}, h)\}_{i}, \quad \mathbf{R}_{h,w}^{front} = \{(w, d_{i})\}_{i}.
\end{equation}
Following the derivation of reference points is the typical practice of deformable attention: we calculate the sampling offsets and attention weights for each reference point through linear layers and sum up the sampled features weighted by their attention score:
\begin{equation}
    \mathrm{CVHA}(\mathbf{t}_{h,w})=\mathrm{DA}(\mathbf{t}_{h,w}, \mathbf{R}_{h,w}, \mathbf{T}).
\end{equation}

\subsection{Applications of TPV}\label{subsec: app}
The TPV planes $\mathbf{T}$ obtained by TPVFormer encode fine-grained view-specific information of a 3D scene.
Still, they are in the form of orthogonal cross-planes and not readily interpretable to common task heads.
Here we explain how to convert TPV planes to point and voxel features and further introduce a lightweight segmentation head.

\textbf{Point Feature.}
Given locations in the real world, we consider the feature generation process as the points querying their features from the TPV representation.
As defined in \eqref{eqn: tpv plane sampling} and \eqref{eqn: tpv plane aggregate}, we first project the points onto the TPV planes to retrieve the corresponding features $[\mathbf{t}_{h,w},\mathbf{t}_{d,h},\mathbf{t}_{w,d}]$, and sum them up to obtain the per-point features.

\textbf{Voxel Feature.}
For dense voxel features, we actively broadcast each TPV plane along the corresponding orthogonal direction to produce three feature tensors of the same size $H\times W\times D\times C$, and aggregate them by summation to obtain the full-scale voxel features.
Note that we do not know the position of any physical point in advance.

To conduct fine-grained segmentation tasks, we apply a lightweight MLP on the point or voxel features to predict their semantic labels, which is instantiated by only two linear layers and an intermediate activation layer.

\definecolor{LightGrey}{rgb}{.9,.9,.9}
\definecolor{White}{rgb}{1.,0.,1.}
\definecolor{first}{rgb}{.8,.0,.0}
\definecolor{second}{rgb}{.0,.6,.0}
\definecolor{third}{rgb}{.0,.0,.8}

\definecolor{nbarrier}{RGB}{255, 120, 50}
\definecolor{nbicycle}{RGB}{255, 192, 203}
\definecolor{nbus}{RGB}{255, 255, 0}
\definecolor{ncar}{RGB}{0, 150, 245}
\definecolor{nconstruct}{RGB}{0, 255, 255}
\definecolor{nmotor}{RGB}{200, 180, 0}
\definecolor{npedestrian}{RGB}{255, 0, 0}
\definecolor{ntraffic}{RGB}{255, 240, 150}
\definecolor{ntrailer}{RGB}{135, 60, 0}
\definecolor{ntruck}{RGB}{160, 32, 240}
\definecolor{ndriveable}{RGB}{255, 0, 255}
\definecolor{nother}{RGB}{139, 137, 137}
\definecolor{nsidewalk}{RGB}{75, 0, 75}
\definecolor{nterrain}{RGB}{150, 240, 80}
\definecolor{nmanmade}{RGB}{213, 213, 213}
\definecolor{nvegetation}{RGB}{0, 175, 0}

\definecolor{car}{rgb}{0.39215686, 0.58823529, 0.96078431}
\definecolor{bicycle}{rgb}{0.39215686, 0.90196078, 0.96078431}
\definecolor{motorcycle}{rgb}{0.11764706, 0.23529412, 0.58823529}
\definecolor{truck}{rgb}{0.31372549, 0.11764706, 0.70588235}
\definecolor{other-vehicle}{rgb}{0.39215686, 0.31372549, 0.98039216}
\definecolor{person}{rgb}{1.        , 0.11764706, 0.11764706}
\definecolor{bicyclist}{rgb}{1.        , 0.15686275, 0.78431373}
\definecolor{motorcyclist}{rgb}{0.58823529, 0.11764706, 0.35294118}
\definecolor{road}{rgb}{1.        , 0.        , 1.        }
\definecolor{parking}{rgb}{1.        , 0.58823529, 1.        }
\definecolor{sidewalk}{rgb}{0.29411765, 0.        , 0.29411765}
\definecolor{other-ground}{rgb}{0.68627451, 0.        , 0.29411765}
\definecolor{building}{rgb}{1.        , 0.78431373, 0.        }
\definecolor{fence}{rgb}{1.        , 0.47058824, 0.19607843}
\definecolor{vegetation}{rgb}{0.        , 0.68627451, 0.        }
\definecolor{trunk}{rgb}{0.52941176, 0.23529412, 0.        }
\definecolor{terrain}{rgb}{0.58823529, 0.94117647, 0.31372549}
\definecolor{pole}{rgb}{1.        , 0.94117647, 0.58823529}
\definecolor{traffic-sign}{rgb}{1.        , 0.        , 0.    }   

\makeatletter
\newcommand{\car@semkitfreq}{3.92}
\newcommand{\bicycle@semkitfreq}{0.03}
\newcommand{\motorcycle@semkitfreq}{0.03}
\newcommand{\truck@semkitfreq}{0.16}
\newcommand{\othervehicle@semkitfreq}{0.20}
\newcommand{\person@semkitfreq}{0.07}
\newcommand{\bicyclist@semkitfreq}{0.07}
\newcommand{\motorcyclist@semkitfreq}{0.05}
\newcommand{\road@semkitfreq}{15.30}  %
\newcommand{\parking@semkitfreq}{1.12}
\newcommand{\sidewalk@semkitfreq}{11.13}  %
\newcommand{\otherground@semkitfreq}{0.56}
\newcommand{\building@semkitfreq}{14.1}  %
\newcommand{\fence@semkitfreq}{3.90}
\newcommand{\vegetation@semkitfreq}{39.3}  %
\newcommand{\trunk@semkitfreq}{0.51}
\newcommand{\terrain@semkitfreq}{9.17} %
\newcommand{\pole@semkitfreq}{0.29}
\newcommand{\trafficsign@semkitfreq}{0.08}
\newcommand{\semkitfreq}[1]{{\csname #1@semkitfreq\endcsname}}

\newcommand{\barrier@nuscenesfreq}{11.79}
\newcommand{\bicycle@nuscenesfreq}{0.18}
\newcommand{\bus@nuscenesfreq}{5.83}
\newcommand{\car@nuscenesfreq}{48.27}
\newcommand{\construction@nuscenesfreq}{1.92}
\newcommand{\motorcycle@nuscenesfreq}{0.54}
\newcommand{\pedestrian@nuscenesfreq}{2.93}
\newcommand{\trafficcone@nuscenesfreq}{0.93}
\newcommand{\trailer@nuscenesfreq}{6.22}
\newcommand{\truck@nuscenesfreq}{20.07}
\newcommand{\driveable@nuscenesfreq}{28.64}
\newcommand{\other@nuscenesfreq}{0.77}
\newcommand{\sidewalk@nuscenesfreq}{6.34}
\newcommand{\terrain@nuscenesfreq}{6.35}
\newcommand{\manmade@nuscenesfreq}{16.10}
\newcommand{\vegetation@nuscenesfreq}{11.08}
\newcommand{\nuscenesfreq}[1]{{\csname #1@nuscenesfreq\endcsname}}

\section{Experiments}

\begin{figure*}
    \centering
    \includegraphics[width=\linewidth]{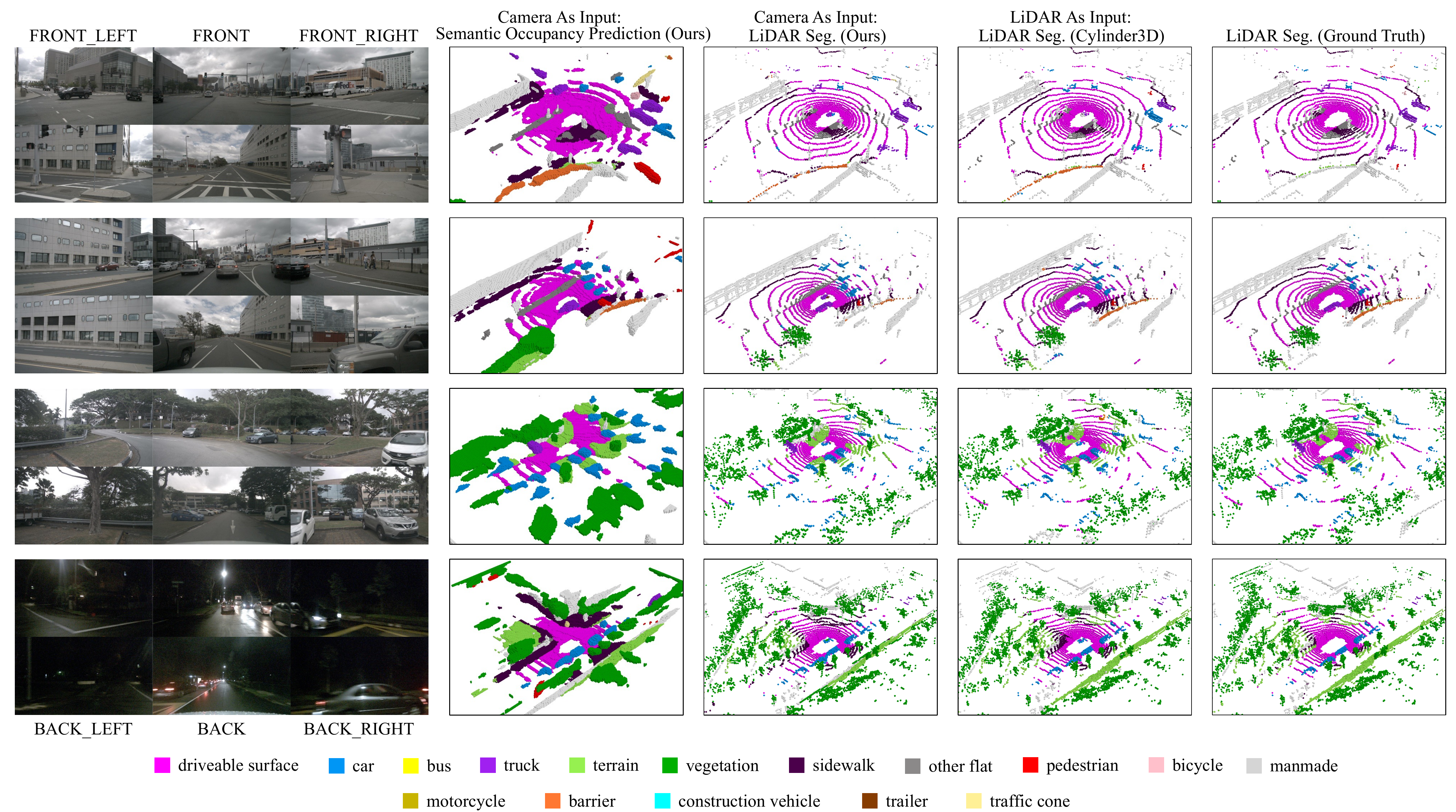}
	\vspace{-7mm}
    \caption{\textbf{Visualization results on 3D semantic occupancy prediction and nuScenes LiDAR segmentation.} 
    Our method can generate more comprehensive prediction results than the LiDAR segmentation ground truth.
    }
    \label{fig: vis main}
	\vspace{-4mm}
\end{figure*}

\begin{figure*}
    \centering
    \includegraphics[width=\linewidth]{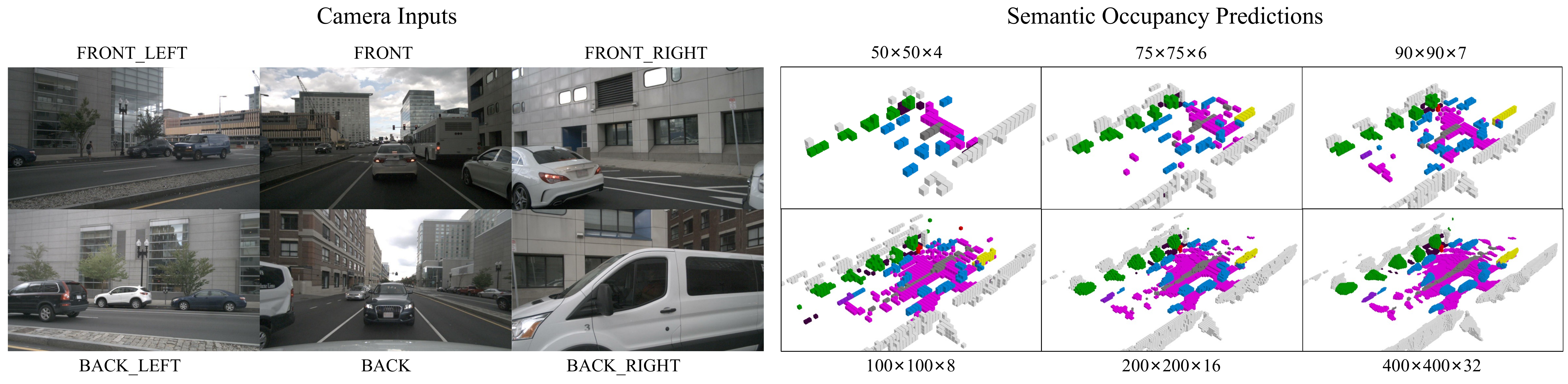}
	\vspace{-7mm}
    \caption{
    \textbf{Arbitrary resolution at test time.}
    We can adjust the prediction resolution through interpolation at test time.
    As resolution increases, more details about the 3D objects are captured.
    }
    \label{fig: vis reso}
	\vspace{-7mm}
\end{figure*}

\begin{table*}[t]
	\footnotesize
	\setlength{\tabcolsep}{0.0045\linewidth}
	\caption{\textbf{LiDAR segmentation results on nuScenes test set.} Despite critical modal difference, our TPVFormer-Base achieves comparable performance with LiDAR-based methods. 
	}
	\vspace{-3mm}
	\newcommand{\classfreq}[1]{{~\tiny(\nuscenesfreq{#1}\%)}}  %
	\centering
	\begin{tabular}{l|c|c | c c c c c c c c c c c c c c c c}
		\toprule
		Method
		& \makecell{Input \\ Modality} & mIoU
		& \rotatebox{90}{\textcolor{nbarrier}{$\blacksquare$} barrier}
		& \rotatebox{90}{\textcolor{nbicycle}{$\blacksquare$} bicycle}
		& \rotatebox{90}{\textcolor{nbus}{$\blacksquare$} bus}
		& \rotatebox{90}{\textcolor{ncar}{$\blacksquare$} car}
		& \rotatebox{90}{\textcolor{nconstruct}{$\blacksquare$} const. veh.}
		& \rotatebox{90}{\textcolor{nmotor}{$\blacksquare$} motorcycle}
		& \rotatebox{90}{\textcolor{npedestrian}{$\blacksquare$} pedestrian}
		& \rotatebox{90}{\textcolor{ntraffic}{$\blacksquare$} traffic cone}
		& \rotatebox{90}{\textcolor{ntrailer}{$\blacksquare$} trailer}
		& \rotatebox{90}{\textcolor{ntruck}{$\blacksquare$} truck}
		& \rotatebox{90}{\textcolor{ndriveable}{$\blacksquare$} drive. suf.}
		& \rotatebox{90}{\textcolor{nother}{$\blacksquare$} other flat}
		& \rotatebox{90}{\textcolor{nsidewalk}{$\blacksquare$} sidewalk}
		& \rotatebox{90}{\textcolor{nterrain}{$\blacksquare$} terrain}
		& \rotatebox{90}{\textcolor{nmanmade}{$\blacksquare$} manmade}
		& \rotatebox{90}{\textcolor{nvegetation}{$\blacksquare$} vegetation}
		\\
		\midrule
		MINet~\cite{minet} & LiDAR & 56.3 & 54.6 & 8.2 & 62.1 & 76.6 & 23.0 & 58.7 & 37.6 & 34.9 & 61.5 & 46.9 & 93.3 & 56.4 & 63.8 & 64.8 & 79.3 & 78.3  \\
		
		PolarNet~\cite{polarnet} & LiDAR & 69.4 & 72.2 & 16.8 & 77.0 & 86.5 & 51.1 & 69.7 & 64.8 & 54.1 & 69.7 & 63.5 & 96.6 & 67.1 & 77.7 & 72.1 & 87.1 & 84.5  \\
		
		PolarSteam~\cite{polarstream} & LiDAR & 73.4 & 71.4 & 27.8 & 78.1 & 82.0 & 61.3 & 77.8 & 75.1 & 72.4 & 79.6 & 63.7 & 96.0 & 66.5 & 76.9 & 73.0 & 88.5 & 84.8  \\
		
		JS3C-Net~\cite{js3c} & LiDAR & 73.6	& 80.1	& 26.2 & 87.8 & 84.5 & 55.2	& 72.6	& 71.3	& 66.3	& 76.8	& 71.2	& 96.8	& 64.5	& 76.9	& 74.1	& 87.5	& 86.1 \\
		
		AMVNet~\cite{amvnet} & LiDAR & 77.3 & 80.6 & 32.0 & 81.7 & 88.9 & 67.1 & 84.3 & 76.1 & 73.5 & 84.9 & 67.3 & 97.5 & 67.4 & 79.4 & 75.5 & 91.5 & 88.7   \\
		
		SPVNAS~\cite{spvnas} & LiDAR & 77.4  & 80.0 & 30.0 & 91.9 & 90.8 & 64.7 & 79.0 & 75.6 & 70.9 & 81.0 & 74.6 & 97.4 & 69.2 & 80.0 & 76.1 & 89.3 & 87.1  \\
		
		Cylinder3D++~\cite{cylinder3D} & LiDAR & 77.9 & 82.8 & 33.9 & 84.3 & 89.4 & 69.6 & 79.4 & 77.3 & 73.4 & 84.6 & 69.4 & 97.7 & 70.2 & 80.3 & 75.5 & 90.4 & 87.6   \\
		
		AF2S3Net~\cite{af2s3net} & LiDAR & 78.3 & 78.9 & \textbf{52.2} & 89.9 & 84.2 & \textbf{77.4} & 74.3 & 77.3 & 72.0 & 83.9 & 73.8 & 97.1 & 66.5 & 77.5 & 74.0 & 87.7 & 86.8   \\
		
		DRINet++~\cite{drinet++} & LiDAR & 80.4 & \textbf{85.5} & 43.2 & 90.5 & \textbf{92.1} & 64.7 & 86.0 & 83.0 & 73.3 & 83.9 & \textbf{75.8} & 97.0 & \textbf{71.0} & \textbf{81.0} & \textbf{77.7} & 91.6 & \textbf{90.2}   \\
		
		LidarMultiNet~\cite{ye2022lidarmultinet} & LiDAR & \textbf{81.4} & 80.4 & 48.4 & \textbf{94.3} & 90.0 & 71.5 & \textbf{87.2} & \textbf{85.2} & \textbf{80.4} & \textbf{86.9} & 74.8 & \textbf{97.8} & 67.3 & 80.7 & 76.5 & \textbf{92.1} & 89.6   \\
		\midrule

		TPVFormer-Small (ours) & Camera & 59.2  & 65.6 & 15.7 & 75.1 & 80.0 & 45.8 & 43.1 & 44.3 & 26.8 & 72.8 & 55.9 & 92.3 & 53.7 & 61.0 & 59.2 & 79.7 & 75.6  \\ %
		
		TPVFormer-Base (ours) & Camera & 69.4  & 74.0 & 27.5 & 86.3 & 85.5 & 60.7 & 68.0 & 62.1 & 49.1 & 81.9 & 68.4 & 94.1 & 59.5 & 66.5 & 63.5 & 83.8 & 79.9  \\ %
		\bottomrule
	\end{tabular}
	\label{tab: main point seg}
	\vspace{-4mm}
\end{table*}

\begin{table*}
	\footnotesize
	\setlength{\tabcolsep}{0.003\linewidth}
	\caption{\textbf{Semantic scene completion results on SemanticKITTI test set.} For fair comparison, we use the performances of RGB-inferred versions of the first four methods reported in MonoScene~\cite{monoscene}.
	We significantly outperform other methods in both IoU and mIoU, including MonoScene which is based on 3D convolution.}
	\vspace{-3mm}
	\newcommand{\classfreq}[1]{{~\tiny(\semkitfreq{#1}\%)}}  %
	\centering
	\begin{tabular}{l|c|c c | c c c c c c c c c c c c c c c c c c c}
		\toprule
		Method
		& \makecell{Input\\ Modality}
		& \makecell{SC\\ IoU} & \makecell{SSC \\ mIoU}
		& \rotatebox{90}{road}
		\rotatebox{90}{\ \ \ \classfreq{road}} 
		& \rotatebox{90}{sidewalk}
		\rotatebox{90}{\ \ \ \classfreq{sidewalk}}
		& \rotatebox{90}{parking}
		\rotatebox{90}{\ \ \ \classfreq{parking}} 
		& \rotatebox{90}{other-grnd}
		\rotatebox{90}{\ \ \ \classfreq{otherground}} 
		& \rotatebox{90}{ building}
		\rotatebox{90}{\ \ \ \classfreq{building}} 
		& \rotatebox{90}{ car}
		\rotatebox{90}{\ \ \ \classfreq{car}} 
		& \rotatebox{90}{ truck}
		\rotatebox{90}{\ \ \ \classfreq{truck}} 
		& \rotatebox{90}{ bicycle}
		\rotatebox{90}{\ \ \ \classfreq{bicycle}} 
		& \rotatebox{90}{motorcycle}
		\rotatebox{90}{\ \ \ \classfreq{motorcycle}} 
		& \rotatebox{90}{ other-veh.}
		\rotatebox{90}{\ \ \  \classfreq{othervehicle}} 
		& \rotatebox{90}{vegetation}
		\rotatebox{90}{\ \ \ \classfreq{vegetation}} 
		& \rotatebox{90}{ trunk}
		\rotatebox{90}{\ \ \ \classfreq{trunk}} 
		& \rotatebox{90}{terrain}
		\rotatebox{90}{\ \ \ \classfreq{terrain}} 
		& \rotatebox{90}{ person}
		\rotatebox{90}{\ \ \ \classfreq{person}} 
		& \rotatebox{90}{ bicyclist}
		\rotatebox{90}{\ \ \ \classfreq{bicyclist}} 
		& \rotatebox{90}{ motorcyclist.}
		\rotatebox{90}{\ \ \ \classfreq{motorcyclist}} 
		& \rotatebox{90}{ fence}
		\rotatebox{90}{\ \ \ \classfreq{fence}} 
		& \rotatebox{90}{ pole}
		\rotatebox{90}{\ \ \ \classfreq{pole}} 
		& \rotatebox{90}{traf.-sign}
		\rotatebox{90}{\ \ \ \classfreq{trafficsign}} 
		\\
		\midrule
		LMSCNet~\cite{lmscnet} & Camera &  31.38 & 7.07 &  46.70 & 19.50 & 13.50 & 3.10 & 10.30 & 14.30 & 0.30 & 0.00 & 0.00 & 0.00 & 10.80 & 0.00 & 10.40 & 0.00 & 0.00 & 0.00 & 5.40 & 0.00 & 0.00   \\
		
		3DSketch~\cite{dsketch} & Camera & 26.85 & 6.23 & 37.70 & 19.80 & 0.00 & 0.00 & 12.10 & 17.10 & 0.00 & 0.00 & 0.00 & 0.00 & 12.10 & 0.00 & 16.10 & 0.00 & 0.00 & 0.00 & 3.40 & 0.00 & 0.00  \\
		
		AICNet~\cite{aicnet} & Camera & 23.93 & 7.09	& 39.30	& 18.30 & 19.80 & 1.60 & 9.60	& 15.30	& 0.70	& 0.00	& 0.00	& 0.00	& 9.60	& 1.90	& 13.50	& 0.00	& 0.00	& 0.00	& 5.00	& 0.10	& 0.00 \\
		
		JS3C-Net~\cite{js3c} & Camera & 34.00 & 8.97 & 47.30 & 21.70 & 19.90 & 2.80 & 12.70 & \textbf{20.10} & 0.80 & 0.00 & 0.00 & \underline{4.10} & \underline{14.20} & \textbf{3.10} & 12.40 & 0.00 & 0.20 & 0.20 & 8.70 & 1.90 & 0.30  \\
		
		MonoScene~\cite{monoscene} & Camera & \underline{34.16} & \underline{11.08} & \underline{54.70} & \underline{27.10} & \underline{24.80} & \underline{5.70} & \underline{14.40} & 18.80 & \underline{3.30} & \underline{0.50} & \textbf{0.70} & \textbf{4.40} & \textbf{14.90} & 2.40 & \underline{19.50} & \underline{1.00} & \underline{1.40} & \textbf{0.40} & \textbf{11.10} & \textbf{3.30} & \textbf{2.10}  \\
		
		TPVFormer (ours) & Camera & \textbf{34.25} & \textbf{11.26} & \textbf{55.10} & \textbf{27.20} & \textbf{27.40} & \textbf{6.50} & \textbf{14.80} & \underline{19.20} & \textbf{3.70} & \textbf{1.00} & \underline{0.50} & 2.30 & 13.90 & \underline{2.60} & \textbf{20.40} & \textbf{1.10} & \textbf{2.40} & \underline{0.30} & \underline{11.00} & \underline{2.90} & \underline{1.50}  \\
		\bottomrule
	\end{tabular}
	\label{tab: main ssc}
	\vspace{-7mm}
\end{table*}

\subsection{Task Descriptions}
We conduct three types of experiments, including 3D semantic occupancy prediction, LiDAR segmentation, and semantic scene completion (SSC). 
The first two tasks are performed on Panoptic nuScenes~\cite{fong2021panoptic}, and the last one is on Semantic KITTI~\cite{semantickitti}.
We detail the datasets in Section \ref{dataset}.
For all tasks, our model only uses RGB images as inputs.

\textbf{3D semantic occupancy prediction.}
As dense semantic labels are difficult to obtain, we formulate a practical yet challenging task for vision-based 3D semantic occupancy prediction.
Under this task, the model is only trained using sparse semantic labels (LiDAR points) but is required to produce a semantic occupancy for all the voxels in the concerned 3D space during testing.
As no benchmark is provided for this, we only perform a qualitative analysis of our method.
Still, our method is the first to demonstrate effective results on this challenging task.

\textbf{LiDAR segmentation.}
The LiDAR segmentation task corresponds to the point querying formulation discussed in Section~\ref{subsec: app}, where we predict the semantic label of a given point.
The LiDAR segmentation task does not necessarily use point clouds as input.
In our case, we use only RGB images as input, while the points are merely used to query their features and for supervision in the training phase.

\textbf{Semantic Scene Completion.}
In conventional SSC, given a single initial LiDAR scan, one needs to predict whether each voxel is occupied and its semantic label for the complete scene inside a certain volume.
As a vision-centric adaptation, we use as input only RGB images and predict the occupancy and semantic label of each voxel.
Accordingly, we supervise the training process with voxel labels.
In the case of TPV representation, we adopt the voxel feature formulation in Section~\ref{subsec: app} to generate full-scale voxel features.
Following common practices, we report the intersection over union (IoU) of occupied voxels, ignoring their semantic class, for the scene completion (SC) task and the mIoU of all semantic classes for the SSC task.

\subsection{Implementation Details}

\textbf{3D semantic occupancy prediction and LiDAR segmentation.}
We construct two versions of TPVFormer, namely TPVFormer-Base and TPVFormer-Small, for different trade-offs between performance and efficiency.
TPVFormer-Base uses the ResNet101-DCN~\cite{resnet,dcn} initialized from FCOS3D~\cite{fcos3d} checkpoint, while TPVFormer-Small adopts the ResNet-50~\cite{resnet} pretrained on ImageNet~\cite{deng2009imagenet}.
Following Cylinder3D~\cite{cylinder3D}, we employ both cross entropy loss and lovasz-softmax~\cite{lovasz} loss to optimize our network.
For lovasz-softmax loss, we use features of real points from LiDAR scans as input to maximize the IoU score for classes, while voxel features are used in cross entropy loss to improve point classification accuracy and avoid semantic ambiguity.
For 3D semantic occupancy prediction, we generate pseudo-per-voxel labels from sparse point cloud by assigning a new label of empty to any voxel that does not contain any point, and we use voxel predictions as input to both lovasz-softmax and cross-entropy losses.

\textbf{Semantic Scene Completion.}
We follow the settings of MonoScene~\cite{monoscene} in the SSC task for fair comparisons.
For model architecture, we adopt the 2D UNet based on a pretrained EfficientNetB7~\cite{efficientnet} as 2D backbone to generate multi-scale image features, which is the same as MonoScene.
For optimization, we employ the losses in MonoScene except for the relation loss.

We provide more details in Section~\ref{app:implement}.

\subsection{3D Semantic Occupancy Prediction Results}

\textbf{Main results.}
In Figure~\ref{fig: vis main}, we provide the main visualization results for SOP.
Our result is much denser and more realistic than the LiDAR segmentation ground truth, which validates the effectiveness of TPV representation in modeling the 3D scene and semantic occupancy prediction.
Furthermore, only querying the LiDAR points results in very close predictions to the ground truth and excels the Cylinder3D counterpart in some cases.
For example, Cylinder3D fails to predict one of the two trucks on the rightmost side of the first scene, while our TPVFormer predicts correctly.

\textbf{Arbitrary resolution at test time.}
Given the simplicity of our segmentation head, we can adjust the resolution of TPV planes at test time arbitrarily without retraining the network.
Figure~\ref{fig: vis reso} shows the results for resolution adjustment, in which we gradually increase the resolution of TPV planes from an initial 50x50x4 to 8 times larger.
It is evident that as resolution increases, TPV representation is able to capture more details about the 3D objects, such as shape.

\textbf{More visualizations} are included in Section~\ref{app:occupancy}.

\subsection{LiDAR segmentation Results}
As the first vision-based method for LiDAR segmentation task, we benchmark TPVFormer against LiDAR-based methods.
As shown in Table~\ref{tab: main point seg}, TPVFormer achieves comparable mIoU ($\sim 70\%$) with most LiDAR-based methods.
This is nontrivial since our method needs to reconstruct the complete 3D scene at a high resolution from only 2D image input, while the 3D structural information is readily available in the point clouds for LiDAR-based methods.
We include results on the validation set in Section~\ref{app:LidarSeg}.

\begin{table} \small
	\caption{\textbf{Different prediction types as input to loss functions for LiDAR segmentation.} Voxel and point in the loss column represent voxel and point predictions. We report mIoUs calculated with both voxel and point predictions.}
	\vspace{-3mm}
		\centering
		\begin{tabular}[b]{cc|cc}
			\toprule
			\multicolumn{2}{c|}{Loss} & \multicolumn{2}{c}{mIoU}
			\\
			CE. & Lovasz & Voxel & Point  
			\\
			\midrule
			Voxel & Voxel & 63.17 & 50.66
			\\
			Voxel & Point & 63.37 & \textbf{64.80}
			\\
			Point & Voxel & \textbf{64.07} & 64.46
			\\
			Point & Point & 49.94 & 64.02
            \\
			\bottomrule
		\end{tabular}
	\label{tab:ablate_loss}
    \vspace{-4mm}
\end{table}

\begin{table}[t] \small
    \setlength{\tabcolsep}{0.01\linewidth}
    \caption{\textbf{Ablations on resolutions and feature dimensions.}}
	\vspace{-3mm}
    \centering
    \begin{tabular}{c|cc|c}
		\toprule
		Method & Resolution & Feature & Point mIoU  \\
		\midrule
		\multirow{2}*{BEVFormer} & 100x100 & 256 & 50.37 \\
		                         & 200x200 & 256 & 56.21 \\
		\midrule
		\multirow{2}*{TPVFormer} & 100x100x8 & 256 & 64.15\\
		                         & 200x200x16 & 128 & \textbf{68.86} \\
		\bottomrule
	\end{tabular}
    \label{tab: ablation other}
    \vspace{-7mm}
\end{table}

\subsection{Semantic Scene Completion Results}
In Table~\ref{tab: main ssc}, we report the results of the semantic scene completion task on SemanticKITTI test set.
We compare our TPVFormer against MonoScene~\cite{monoscene}, which is a vision-based method based on 3D convolution in the voxel space.
We also include the 4 baseline methods provided in MonoScene~\cite{monoscene}.
TPVFormer outperforms all other methods in both IoU and mIoU, which demonstrates the effectiveness of TPVFormer in occupancy and semantics prediction.
Furthermore, TPVFormer enjoys significant advantages over MonoScene in both parameter number and computation.
Specifically, TPVFormer has only 6.0M parameters versus 15.7M for MonoScene, and 128G FLOPS per image versus 500G for MonoScene.
We report results on the validation set in Section~\ref{app:SSC}.

\subsection{Abation Study}
We ablate our TPVFormer on the validation sets of nuScenes and SemanticKITTI for LiDAR segmentation and semantic scene completion, respectively.

\textbf{Loss functions for LiDAR segmentation.}
We employ both cross entropy (CE.) loss and lovasz-softmax loss~\cite{lovasz} for LiDAR segmentation.
As our TPVFormer can produce point-level and voxel-level predictions in a single forward propagation, we investigate different prediction types as input to these loss functions.
As shown in Table~\ref{tab:ablate_loss}, when both voxel and point predictions are used as input to the loss functions, the mIoUs from both predictions are high and close to each other.
However, when only voxel or point prediction is employed in optimization, the corresponding mIoU will be much higher than the other one.
We think the TPV representation might be discretized in the voxel space if given only voxel-level supervision, and thus the interpolation used to generate point predictions will not apply.
In cases with only point-level supervision, TPV fails to learn the discretization strategy implied in the voxel space.

\begin{table}[t] \small
    \caption{\textbf{Different number of HCAB blocks and HAB blocks for semantic scene completion.} We keep the total number of attention modules the same in these experiments.}
	\vspace{-3mm}
    \centering
    \begin{tabular}{cc|cc}
			\toprule
			\# HCAB  & \# HAB & SC IoU & SSC mIoU  
			\\
			\midrule
			2 & 4 & 35.55 & 10.49
			\\
			3 & 2 & 35.61 & \textbf{11.36}
			\\
			4 & 0 & \textbf{35.79} & 10.82
			\\
			\bottomrule
		\end{tabular}
    \label{tab: ablation blocks}
    \vspace{-7mm}
\end{table}

\textbf{Other design choices.}
We ablate other design choices and compare our method with an adaptation of BEVFormer in Table~\ref{tab: ablation other}.
TPVFormer favors resolution more than feature dimension because increasing the resolution is a direct way to enhance its ability for modelling more fine-grained structures.
In addition, our method performs better than BEVFormer under all configurations, which confirms that TPV substantially improves the ability to describe fine-grained structures with three complementary cross-views.

\textbf{The number of HCAB and HAB blocks.}
As HCAB and HAB blocks aggregate visual information from image features and contextual information from other TPV queries, respectively, we study the proportion of the two blocks in Table~\ref{tab: ablation blocks}.
The IoU improves as increasing the number of HCAB blocks, validating the importance of direct visual clues for geometry understanding.
However, the semantic prediction relies on both visual and contextual information as the maximum mIoU is achieved with a moderate number of HCAB and HAB blocks.

\section{Conclusion}
In this paper, we have presented a tri-perspective view (TPV) representation, which is able to describe the fine-grained structures of a 3D scene efficiently.
To lift image features to the 3D TPV space, we have proposed a TPVFormer model based on the attention mechanism.
The visualization results have shown that our TPVFormer produces consistent semantic voxel occupancy prediction with only sparse point supervision during training.
We have demonstrated for the first time that our vision-based method achieves comparable performance with LiDAR-based methods on nuScenes LiDAR segmentation task.

\appendix

\begin{figure*}[t]
\centering
\includegraphics[width=\textwidth]{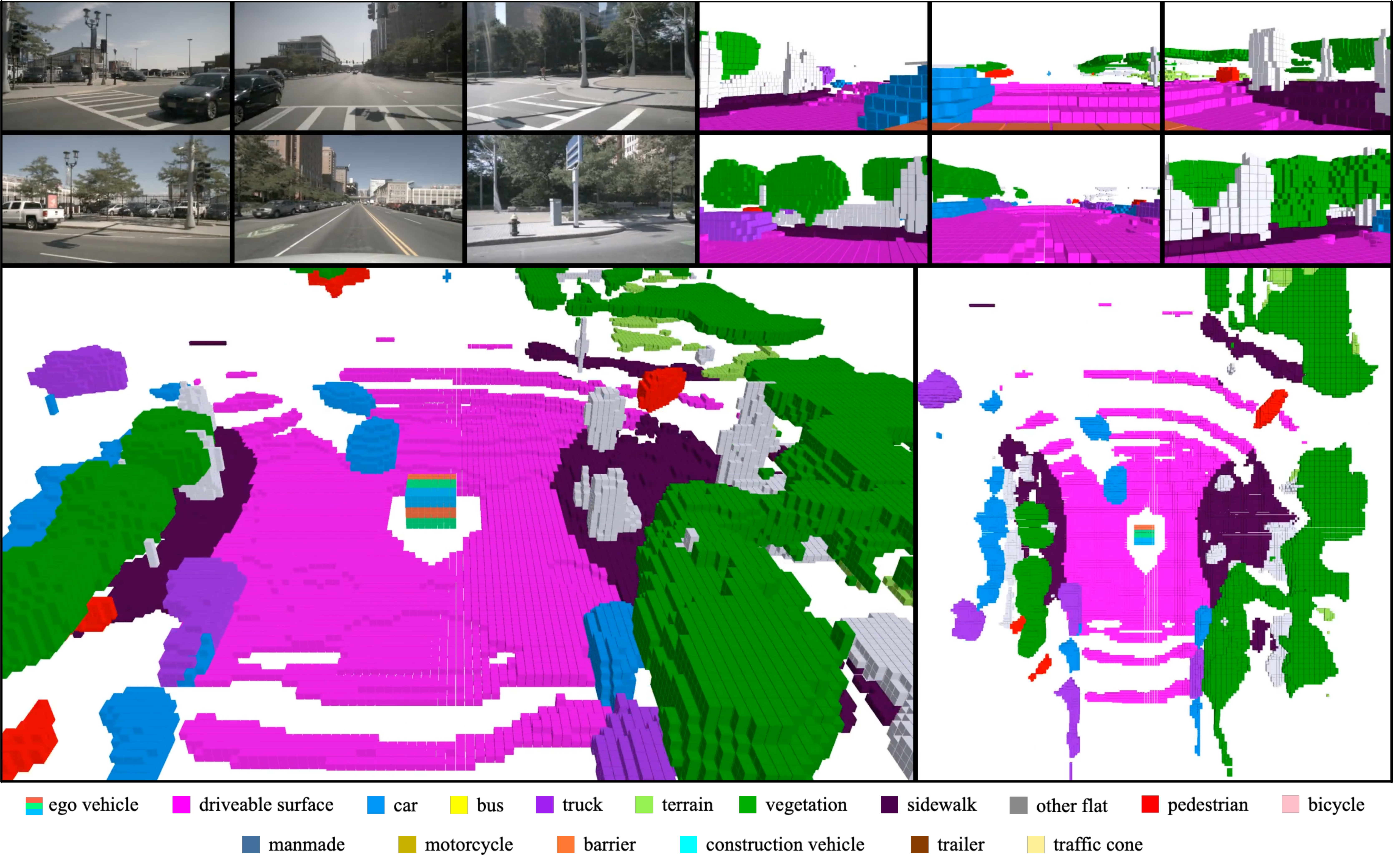}
\vspace{-7mm}
\caption{An image sampled from the video demo for 3D semantic occupancy prediction on nuScenes validation set (not seen in the training phase). 
    We predict the semantic occupancies for all voxels in the 3D space.
    The six images in the top left are the inputs to our model captured by the front-left, front, front-right, back-left, back, and back-right cameras. 
    The six images in the top right denote our prediction results with the corresponding views as the inputs.
    The bottom two images provide a global view of our predictions where the red-green-blue box represents the ego vehicle.
}
\label{teaser_supp}
\vspace{-5mm}
\end{figure*}

\appendix

\section{Dataset Details} \label{dataset}

\textbf{The Panoptic nuScenes dataset}~\cite{fong2021panoptic} collects 1000 driving scenes of 20 seconds duration each, and the keyframes are annotated at 2Hz.
Each sample contains RGB images from 6 cameras with $360^{\circ}$ horizontal FOV and point cloud data from 32 beams LiDAR sensor.
The total of 1000 scenes are officially divided into training, validation and test splits with 700, 150 and 150 scenes, respectively.

\textbf{The SemanticKITTI dataset}~\cite{semantickitti} is a large-scale outdoor-scene dataset, which includes automotive LiDAR scans voxelized into $256\times 256\times 32$ grids.
Each voxel has a side length of 0.2m and is labeled with one of 21 classes (19 semantic, 1 free and 1 unknown).
In our experiments, we also use RGB images captured by cam2 from the KITTI odometry benchmark.
The voxel and image data is officially arranged as 22 sequences, split into 10/1/11 sequences for training, validation and test. 

\section{Implementation Details} \label{app:implement}
\textbf{3D semantic occupancy prediction and LiDAR segmentation.}
TPVFormer-Base uses the ResNet101-DCN~\cite{resnet,dcn} initialized from FCOS3D~\cite{fcos3d} checkpoint, while TPVFormer-Small adopts the ResNet-50~\cite{resnet} pretrained on ImageNet~\cite{deng2009imagenet}.
The TPV resolutions are 200x200x16 and 100x100x8 for the base and small versions, respectively, and we upsample the TPV planes by a factor of 2 in TPVFormer-Small for finer supervision.
Although both of them share the same TPV feature dimension of 128, the base model uses multi-scale image features and an input image resolution of 1600x900 instead of single-scale image features and 800x450 input for the small model.

For training, we adopt the AdamW~\cite{adamw} optimizer with initial learning rate as 2e-4 and weight decay as 0.01.
We use the cosine learning rate scheduler with a linear warming up in the first 500 iterations, and the same image augmentation strategy as BEVFormer~\cite{bevformer}.
All models are trained for 24 epochs with a batch size of 8 on 8 A100 GPUs.

\textbf{Semantic Scene Completion.}
We adopt the 2D UNet based on a pretrained EfficientNetB7~\cite{efficientnet} as 2D backbone to generate multi-scale image features, which is the same as MonoScene.
Moreover, we set the resolution of TPV planes as 128x128x16 to generate a 3D voxel feature tensor of the same size as MonoScene, although our TPV planes are 2D feature maps while MonoScene operates directly on 3D voxel features.
We use RGB images from cam2 cropped to 1220x370 as input and a feature dimension of 96.
For optimization, we employ the losses in MonoScene except for the relation loss, since TPVFormer does not have the 3D CRP module or any downsampling operation.
For training, we generally follow the recipe in MonoScene.
Specifically, we use a learning rate of 2e-4, a weight decay of 0.01, and a cosine scheduler.
We keep the other settings the same.
For a fair comparison, we also rerun the official code of MonoScene with a cosine learning rate scheduler.

\section{3D Semantic Occupancy Prediction Results} \label{app:occupancy}
\begin{figure}[t]
\centering
\includegraphics[width=0.475\textwidth]{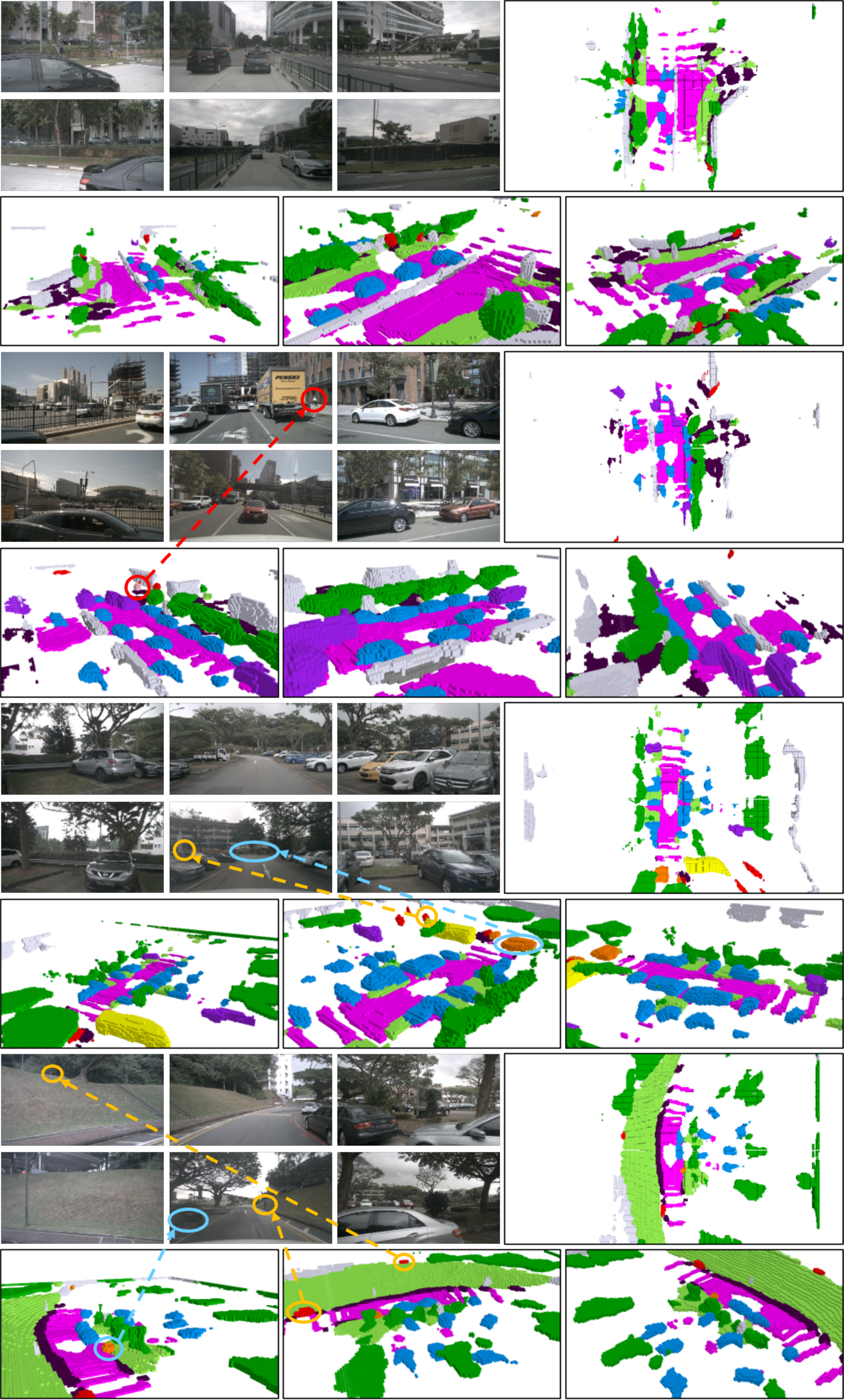}
\vspace{-7mm}
\caption{More visualizations of the proposed TPVFormer for 3D semantic occupancy prediction.
}
\label{fig: supp details}
\vspace{-5mm}
\end{figure}

We provide a video demo~\footnote{\url{https://github.com/wzzheng/TPVFormer}.} for 3D semantic occupancy prediction on nuScenes validation set with a sampled image in Figure~\ref{teaser_supp}.
Figure~\ref{fig: supp details} provides detailed visualization results of our model for four samples from nuScenes validation set.
For each sample, we present the six surround camera images, the top view of the predicted scene, and the zoomed-in results from three different angles. 
In addition, we highlight predictions for small and rare objects with circles and further link them to corresponding ground truths in RGB images with arrowed dash lines.
Specifically, we highlight bicycles, motorcycles and pedestrians with red, blue and yellow circles, respectively.
Note that although some of these objects are barely visible in RGB images, our model still predicts them successfully.

\section{LiDAR segmentation Results} \label{app:LidarSeg}

\begin{table*}[t]
	\footnotesize
	\setlength{\tabcolsep}{0.0045\linewidth}
	\caption{\textbf{LiDAR segmentation results on nuScenes validation set.} Despite critical modal difference, our TPVFormer-Base achieves comparable performance with LiDAR-based methods. 
	Moreover, the mIoU gap between BEVFormer and TPVFormer clearly proves the effectiveness of TPV in modelling fine-grained 3D structures of a scene.
	}
	\vspace{-2mm}
	\newcommand{\classfreq}[1]{{~\tiny(\nuscenesfreq{#1}\%)}}  %
	\centering
	\begin{tabular}{l|c|c | c c c c c c c c c c c c c c c c}
		\toprule
		Method
		& \makecell{Input \\ Modality} & mIoU
		& \rotatebox{90}{\textcolor{nbarrier}{$\blacksquare$} barrier}
		
		& \rotatebox{90}{\textcolor{nbicycle}{$\blacksquare$} bicycle}
		
		& \rotatebox{90}{\textcolor{nbus}{$\blacksquare$} bus}

		& \rotatebox{90}{\textcolor{ncar}{$\blacksquare$} car}

		& \rotatebox{90}{\textcolor{nconstruct}{$\blacksquare$} const. veh.}

		& \rotatebox{90}{\textcolor{nmotor}{$\blacksquare$} motorcycle}

		& \rotatebox{90}{\textcolor{npedestrian}{$\blacksquare$} pedestrian}

		& \rotatebox{90}{\textcolor{ntraffic}{$\blacksquare$} traffic cone}

		& \rotatebox{90}{\textcolor{ntrailer}{$\blacksquare$} trailer}

		& \rotatebox{90}{\textcolor{ntruck}{$\blacksquare$} truck}

		& \rotatebox{90}{\textcolor{ndriveable}{$\blacksquare$} drive. suf.}

		& \rotatebox{90}{\textcolor{nother}{$\blacksquare$} other flat}

		& \rotatebox{90}{\textcolor{nsidewalk}{$\blacksquare$} sidewalk}

		& \rotatebox{90}{\textcolor{nterrain}{$\blacksquare$} terrain}

		& \rotatebox{90}{\textcolor{nmanmade}{$\blacksquare$} manmade}

		& \rotatebox{90}{\textcolor{nvegetation}{$\blacksquare$} vegetation}

		\\
		\midrule

        RangeNet++~\cite{rangenet++} & LiDAR &  65.5 & 66.0 & 21.3 & 77.2 & 80.9 & 30.2 & 66.8 & 69.6 &  52.1 & 54.2 & {72.3} & {94.1} & 66.6 & 63.5 & 70.1 & 83.1 & 79.8 \\
		
		PolarNet~\cite{polarnet} & LiDAR & 71.0 & 74.7 & 28.2 & 85.3 & 90.9 & 35.1 & 77.5 & 71.3 & 58.8 & 57.4 & 76.1 & 96.5 & 71.1 & 74.7 & 74.0 & 87.3 & 85.7  \\
		
		Salsanext~\cite{salsanext} & LiDAR & 72.2 & 74.8 & 34.1 & 85.9 & 88.4 & 42.2 & 72.4 & 72.2 & 63.1 & 61.3 & 76.5 & 96.0 & 70.8 & 71.2 & 71.5 & 86.7 & 84.4 \\

		Cylinder3D++~\cite{cylinder3D} & LiDAR & \bf{76.1} & \bf{76.4} & \bf{40.3} & \bf{91.2} & \bf{93.8} & \textbf{51.3} & \bf{78.0} & \bf{78.9} & \bf{64.9} & \bf{62.1} & \bf{84.4} & \bf{96.8} & \bf{71.6} & \bf{76.4} & \bf{75.4} & \bf{90.5} & \bf{87.4}  \\
			\midrule

        BEVFormer-Base~\cite{bevformer} & Camera & 56.2  & 54.0 & 22.8 & 76.7 & 74.0 & 45.8 & 53.1 & 44.5 & 24.7 & 54.7 & 65.5 & 88.5 & 58.1 & 50.5 & 52.8 & 71.0 & 63.0  \\

		TPVFormer-Small (ours) & Camera & 59.3  & 64.9 & 27.0 & 83.0 & 82.8 & 38.3 & 27.4 & 44.9 & 24.0 & 55.4 & 73.6 & 91.7 & 60.7 & 59.8 & 61.1 & 78.2 & 76.5  \\ %
		
		TPVFormer-Base (ours) & Camera & 68.9  & 70.0 & 40.9 & 93.7 & 85.6 & 49.8 & 68.4 & 59.7 & 38.2 & 65.3 & 83.0 & 93.3 & 64.4 & 64.3 & 64.5 & 81.6 & 79.3  \\ %
		\bottomrule
	\end{tabular}
	\label{tab: supp lidar seg}
	\vspace{-3mm}
\end{table*}

In Table~\ref{tab: supp lidar seg}, we report the performance of TPVFormer on nuScenes validation set for LiDAR segmentation.
For a fair comparison, we replace the temporal module in BEVFormer with self-attention moduel and use a feature dimension of 256 to make the model sizes of BEVFormer-Base and TPVFormer-Base comparable.
The mIoU of TPVFormer-Base is on par with LiDAR-based methods despite critical modal differences.
Furthermore, our TPVFormer-Base achieves a 12.7\% higher mIoU than BEVFormer-Base, which demonstrates the effectiveness of TPV in modeling fine-grained 3D structures of a scene.

\section{Semantic Scene Completion Results} \label{app:SSC}
We present the semantic scene completion performance on SemanticKITTI validation set in Table~\ref{tab: supp ssc}.
Although TPVFormer does not achieve the highest IoU for scene completion, it outperforms other methods in mIoU with a clear margin for semantic scene completion.
We reproduce MonoScene~\cite{monoscene} with the official code in our environment and also report its performance using the cosine learning rate following our recipe for a fair comparison.

\begin{table*}
	\footnotesize
	\setlength{\tabcolsep}{0.003\linewidth}
	\caption{\textbf{Semantic scene completion results on SemanticKITTI validation set.} For a fair comparison, we use the performances of RGB-inferred versions of the first four methods reported in MonoScene~\cite{monoscene}.
	$^*$ represents the reproduced result using the official code.
	$^{**}$ represents result using the cosine learning rate schedule.
	}
	\vspace{-2mm}
	\newcommand{\classfreq}[1]{{~\tiny(\semkitfreq{#1}\%)}}  %
	\centering
	\begin{tabular}{l|c|c c | c c c c c c c c c c c c c c c c c c c}
		\toprule
		Method
		& \makecell{Input\\ Modality}
		& \makecell{SC\\ IoU} & \makecell{SSC \\ mIoU}
		& \rotatebox{90}{road}
		\rotatebox{90}{\classfreq{road}} 
		& \rotatebox{90}{sidewalk}
		\rotatebox{90}{\classfreq{sidewalk}}
		& \rotatebox{90}{parking}
		\rotatebox{90}{\classfreq{parking}} 
		& \rotatebox{90}{other-grnd}
		\rotatebox{90}{\classfreq{otherground}} 
		& \rotatebox{90}{ building}
		\rotatebox{90}{\classfreq{building}} 
		& \rotatebox{90}{ car}
		\rotatebox{90}{\classfreq{car}} 
		& \rotatebox{90}{ truck}
		\rotatebox{90}{\classfreq{truck}} 
		& \rotatebox{90}{ bicycle}
		\rotatebox{90}{\classfreq{bicycle}} 
		& \rotatebox{90}{motorcycle}
		\rotatebox{90}{\classfreq{motorcycle}} 
		& \rotatebox{90}{ other-veh.}
		\rotatebox{90}{\classfreq{othervehicle}} 
		& \rotatebox{90}{vegetation}
		\rotatebox{90}{\classfreq{vegetation}} 
		& \rotatebox{90}{ trunk}
		\rotatebox{90}{\classfreq{trunk}} 
		& \rotatebox{90}{terrain}
		\rotatebox{90}{\classfreq{terrain}} 
		& \rotatebox{90}{ person}
		\rotatebox{90}{\classfreq{person}} 
		& \rotatebox{90}{ bicyclist}
		\rotatebox{90}{\classfreq{bicyclist}} 
		& \rotatebox{90}{ motorcyclist.}
		\rotatebox{90}{\classfreq{motorcyclist}} 
		& \rotatebox{90}{ fence}
		\rotatebox{90}{\classfreq{fence}} 
		& \rotatebox{90}{ pole}
		\rotatebox{90}{\classfreq{pole}} 
		& \rotatebox{90}{traf.-sign}
		\rotatebox{90}{\classfreq{trafficsign}} 
		\\
		\midrule
		LMSCNet~\cite{lmscnet} & Camera & 28.61 & 6.70 & 40.68 & 18.22 & 4.38 & 0.00 & 10.31 & 18.33 & 0.00 & 0.00 & 0.00 & 0.00 & 13.66 & 0.02 & 20.54 & 0.00 & 0.00 & 0.00 & 1.21 & 0.00 & 0.00   \\
		
		3DSketch~\cite{dsketch} & Camera & 33.30 & 7.50 & 41.32 & 21.63 & 0.00 & 0.00 & \underline{14.81} & 18.59 & 0.00 & 0.00 & 0.00 & 0.00 & \textbf{19.09} & 0.00 & 26.40 & 0.00 & 0.00 & 0.00 & 0.73 & 0.00 & 0.00  \\
		
		AICNet~\cite{aicnet} & Camera & 29.59 & 8.31 & 43.55 & 20.55 & {11.97} & {0.07} & 12.94 & 14.71 & {4.53} & 0.00 & 0.00 & 0.00 & 15.37 & \underline{2.90} & {28.71} & 0.00 & 0.00 & 0.00 & 2.52 & 0.06 & 0.00 \\
		
		JS3C-Net~\cite{js3c} & Camera & \textbf{38.98} & 10.31 & 50.49 & {23.74} & 11.94 & {0.07} & \textbf{15.03} & \textbf{24.65} & 4.41 & 0.00 & 0.00 & \textbf{6.15} & \underline{18.11} & \textbf{4.33} & 26.86 & {0.67} & 0.27 & 0.00 & {3.94} & {3.77} & {1.45}  \\
		
		MonoScene$^*$~\cite{monoscene} & Camera & \underline{36.86} & \underline{11.08} & \textbf{56.52} & \textbf{26.72} & {14.27} & {0.46} & {14.09} & 23.26 & \underline{6.98} & \underline{0.61} & \underline{0.45} & {1.48} & {17.89} & 2.81 & \underline{29.64} & \textbf{1.86} & \textbf{1.20} & {0.00} & {5.84} & \textbf{4.14} & \textbf{2.25}  \\
		
		MonoScene$^{**}$~\cite{monoscene} & Camera & 36.13 & 10.98 & {56.30} & \underline{25.89} & \underline{15.91} & \underline{0.75} & {13.47} & 23.31 & {5.36} & \textbf{0.72} & \textbf{0.91} & {3.77} & {17.70} & 2.45 & {27.12} & \underline{1.71} & \underline{1.08} & {0.00} & \textbf{6.34} & \underline{3.79} & \underline{2.03}  \\
		
		TPVFormer (ours) & Camera & 35.61 & \textbf{11.36} & \underline{56.50} & {25.87} & \textbf{20.60} & \textbf{0.85} & {13.88} & \underline{23.81} & \textbf{8.08} & {0.36} & {0.05} & \underline{4.35} & 16.92 & {2.26} & \textbf{30.38} & {0.51} & {0.89} & {0.00} & \underline{5.94} & {3.14} & {1.52}  \\
		\bottomrule
	\end{tabular}
	\label{tab: supp ssc}
	\vspace{-5mm}
\end{table*}

\end{document}